\documentclass[letterpaper, 10 pt, conference]{ieeeconf}  % Comment this line out if you need a4paper

\IEEEoverridecommandlockouts                              % This command is only needed if 
\usepackage{graphicx} % for pdf, bitmapped graphics files
\usepackage{caption}
\usepackage{subcaption}
\usepackage{multirow}
\usepackage{amsmath} % assumes amsmath package installed
\usepackage{amssymb}  % assumes amsmath package installed

\title{\LARGE \bf
Stochastic Action Prediction for Imitation Learning
}

\author{Sagar Gubbi Venkatesh$^{1, 2}$, Nihesh Rathod$^{1, 2}$, Shishir Kolathaya$^{2}$ and Bharadwaj Amrutur$^{1, 2}$% <-this % stops a space
\thanks{*This work was supported by Yaskawa India and Robert Bosch Center for Cyber Physical Systems}% <-this % stops a space
\thanks{$^{1}$Department of Electrical and Communication Engineering, Indian Institute of Science, Bangalore 560012, India {\tt\small sagar@iisc.ac.in}; {\tt\small nihesh@iisc.ac.in}; {\tt\small amrutur@iisc.ac.in}}%
\thanks{$^{2}$Robert Bosch Center for Cyber Physical Systems, Indian Institute of Science, Bangalore 560012, India {\tt\small shishirk@iisc.ac.in}}%
}

\begin{document}

\maketitle
\thispagestyle{empty}
\pagestyle{empty}

%%%%%%%%%%%%%%%%%%%%%%%%%%%%%%%%%%%%%%%%%%%%%%%%%%%%%%%%%%%%%%%%%%%%%%%%%%%%%%%%
\begin{abstract}

Imitation learning is a data-driven approach to acquiring skills that relies on expert demonstrations to learn a policy that maps observations to actions. When performing demonstrations, experts are not always consistent and might accomplish the same task in slightly different ways. In this paper, we demonstrate inherent stochasticity in demonstrations collected for tasks including line following with a remote-controlled car and manipulation tasks including reaching, pushing, and picking and placing an object. We model stochasticity in the data distribution using autoregressive action generation, generative adversarial nets, and variational prediction and compare the performance of these approaches. We find that accounting for stochasticity in the expert data leads to substantial improvement in the success rate of task completion.

\end{abstract}

%%%%%%%%%%%%%%%%%%%%%%%%%%%%%%%%%%%%%%%%%%%%%%%%%%%%%%%%%%%%%%%%%%%%%%%%%%%%%%%%
\section{INTRODUCTION}

Traditionally robot controllers have been designed by modeling the environment and the sensors and actuators in the robot and then analytically arriving at the controller\cite{analytic}. However, this approach becomes difficult as the complexity of the environment and the robot increase. For example, it is difficult to hand-craft a controller that uses visual feedback for complex manipulation tasks\cite{vrteleop}.

An alternative way to build robot controllers is to use learning based methods. Imitation learning is one such data-driven method that has been successfully applied\cite{dagger} to a wide range of problems ranging from self-driving cars\cite{nvidia} and drone navigation\cite{drone} to complex manipulation tasks with a robotic arm\cite{vrteleop}\cite{daml}\cite{yaskawa}\cite{siamese}. The basic principle in imitation learning is to copy the behavior of an expert performing the task. As the expert performs the task by teleoperating the robot, the observations seen by the expert along with the corresponding actions taken are recorded. A neural network is  trained on the expert data collected to clone the policy of the expert.

%For imitation learning to be successful, good quality expert data is essential. However, even with abundant carefully recorded expert demonstrations, there are several potential problems with imitation learning. One, the behaviour of the expert might be non-markovian and dependent on choices made in the past. Two, a distributional drift arises as the trajectory of the cloned policy deviates over time from the expert policy. Three, the behaviour of the expert is truly random or at least cannot be captured in the data collected. This is aleatoric uncertainty as opposed to epistemic uncertainty which can be explained away given enough data.

%The non-markovian behaviour of the expert can be modeled using recurrent neural networks such as LSTMs\cite{lstm} that keep track of history. Several approaches have been proposed to address distributional drift. For example, one can add additional cameras on either side of the main camera to augment the training dataset for navigation tasks\cite{nvidia}\cite{drone}. A more general method to augment the expert dataset is DAGGER\cite{dagger}, where experts are repeatedly asked what they would do had they been in the same state the the policy network finds itself in. Finally, to the model the "true" randomness in the data or aleatoric uncertainty\cite{bayesian_uncertainty}, deterministic neural networks alone are not sufficient and probabilistic elements will have to be introduced. This issue is the subject of this manuscript.

For imitation learning to be successful, good quality expert data is essential. However, even with abundant carefully recorded demonstrations, the expert performing the demonstrations might act differently given the same observations. In these scenarios, the behaviour of the expert is truly random or at least cannot be captured in the data. This is aleatoric uncertainty as opposed to epistemic uncertainty which can be explained away given enough data. The ``true" randomness\cite{bayesian_uncertainty} in the labels has to modeled rather than ignored as noise in the dataset. %To model the ``true" randomness in the data, deterministic neural networks alone are not sufficient.% and probabilistic elements will have to be introduced.

\begin{figure}[!t]
      \centering
      \includegraphics[width=0.8\linewidth]{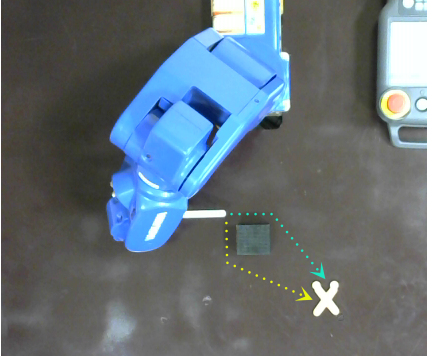}
      \caption{An illustration of randomness in the expert data. The peg may be moved downward or to the right to avoid the obstacle and reach the target at the right-bottom.}
      \label{fig:intro_reach}
\end{figure}

An illustration of randomness in the expert data is shown in Fig~\ref{fig:intro_reach}. The task is to move the peg towards the object at the right-bottom corner while avoiding the obstacle. There are multiple valid ways an expert might behave when faced with the obstacle, and the training dataset might include multiple ways to get around the obstacle. The peg may be moved horizontally towards the right, or the peg may be moved vertically downward when facing the obstacle. If the action space is two dimensional with one of the dimensions for moving horizontally and the other for moving vertically, then the two valid actions are $(RIGHT, HOLD)$ and $(HOLD, DOWN)$. Either of those actions is valid, but sampling the two action dimensions independently could lead to $(RIGHT, DOWN)$ which would cause a collision, even though this is not present in the training dataset.

We set out to answer two questions in this paper:
\begin{itemize}
    \item Is stochasticity prevalent in datasets for imitation learning tasks such as navigation and manipulation tasks like pick-and-place?
    \item If stochasticity is an issue, what approach to modeling it works better for imitation learning?
\end{itemize}

To answer the first question, we built (a) an RC car that can be controlled with a game controller, and (b) a teleoperation system which allows the Yaskawa MotoMini and the Dobot Magician to be controlled by a game controller. We collected demonstrations for line following with the RC car and for manipulation tasks including reaching, pushing, and picking and placing an object with a robot arm.

To address the second question, we compare three prevalent ways to model stochasticity: (a) autoregressive conditional action generation\cite{pixelcnn}\cite{pixelrnn}, (b) generative adversarial networks (GANs)\cite{gan}, and (c) variational action prediction\cite{variational}. We quantify the performance of each method for the navigation and manipulation tasks and discuss the merits and demerits of each method as applicable to imitation learning.

The rest of this paper is organized as follows. In the following section, papers related to this work are discussed. In Section~3, the different potential solutions to modeling stochasticity are discussed. Section~4 details experimental results, and Section~5 concludes the paper.

\section{RELATED WORK}

Deep neural networks have been used to clone expert behaviour where the expert behavior is captured by teleoperation of a robotic arm\cite{vrteleop}. In \cite{mdn}, the authors note that tasks can be accomplished in multiple ways. As a result, the expert data comprising (state, action) pairs could be multi-modal. They find that modeling the multi-modal nature of the expert data with a Gaussian mixture model using a mixture density network substantially improves performance compared to minimizing the mean squared error. We explore a closely related problem of modeling stochasticity when the action space is multi-dimensional.

When generating images, the high dimensionality of the output space makes it intractable to directly model the joint probability of the output. The PixelCNN method described in \cite{pixelcnn} models the joint distribution over the pixels of the output image as a product of conditional distributions where the distribution at each pixel is dependent on the values sampled for the previous pixels in raster scan order. Although \cite{pixelcnn} applies this to images, the same method can be applied wherever the output space is multi-dimensional. In this paper, we apply the PixelCNN approach to policy networks for predicting the action to be taken.

Generative adversarial networks (GANs) have been used to generate realistic images\cite{gan}. In the conditional version of GAN\cite{cgan}, both the generator and discriminator are provided with the data on which we wish to condition. We use conditional GANs for policy networks by conditioning both the generator and the discriminator on the observation and the generator produces actions. One of the major problems with GANs is mode collapse where the generator reaches a local minima where it produces only one or a few modes of the desired distribution and successfully confuses the discriminator. Variants of GANs such as the Wasserstein GAN\cite{wgan} have been proposed to overcome mode collapse. While GANs have most popularly been used for generating images, they can be used as policy networks to generate actions conditioned on observations.

Yet another way to address the problem of stochasticity is with variational prediction. In predicting future frames of a video given the current and past frames, variational video prediction\cite{variationalvid} has been used to address the problem of low quality predictions with significant blurring when deterministic neural networks are used. Variational video prediction has also been leveraged for constructing world models that can then be used for sample efficient model-based reinforcement learning\cite{modelatari}. Variational predictors use the ``reparameterization trick"\cite{variational} when training to refactor stochastic nodes in the network into a differentiable function of its parameters and a random variable from a fixed distribution. In \cite{concrete} and \cite{gumbelsoftmax}, concrete random variables or continuous relaxations of discrete random variables are introduced to reparameterize categorical latent variables. These are used as random latent variables in variational inference\cite{modelatari}. Similar to GANs, variational prediction can be used for policy networks to predict actions given the observation.

\section{PRELIMINARIES}

\begin{figure}[tp]
      \centering
      \includegraphics[width=0.95\linewidth]{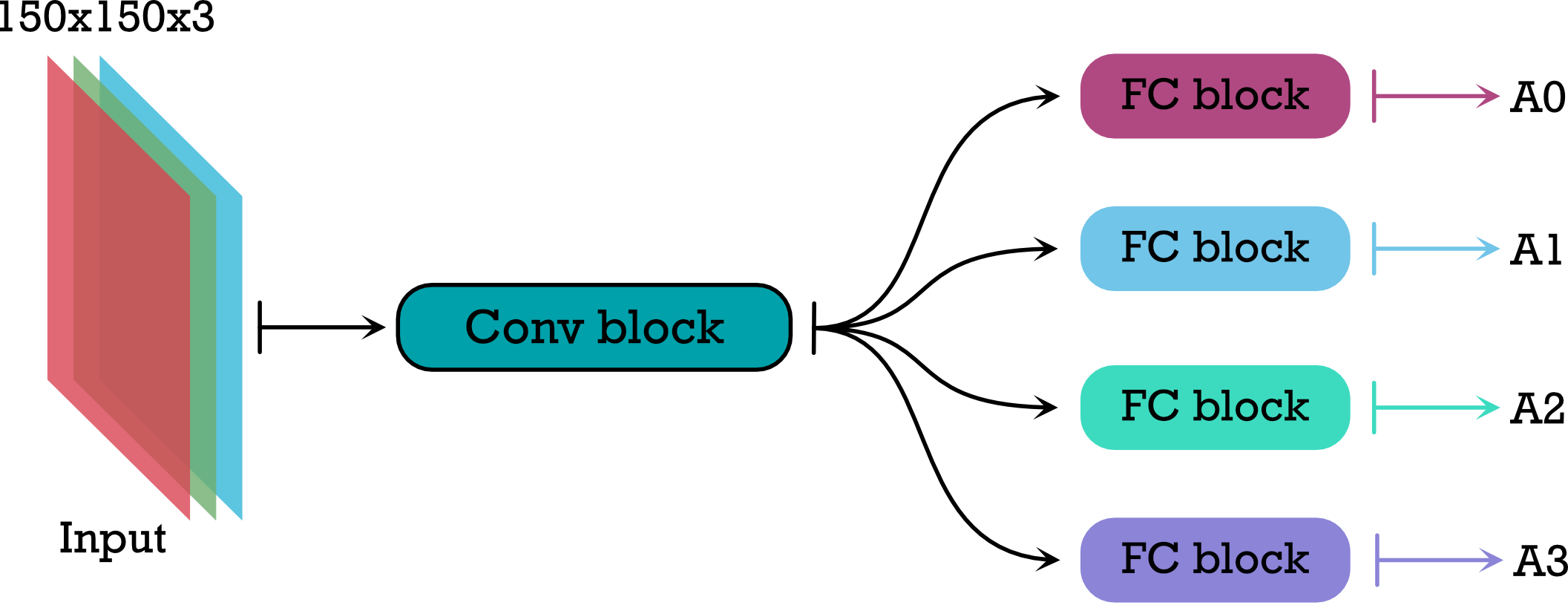}
      \caption{Independently sampling actions.}
      \label{fig:independent_conv}
\end{figure}

\begin{figure}[!t]
      \centering
      \includegraphics[width=0.95\linewidth]{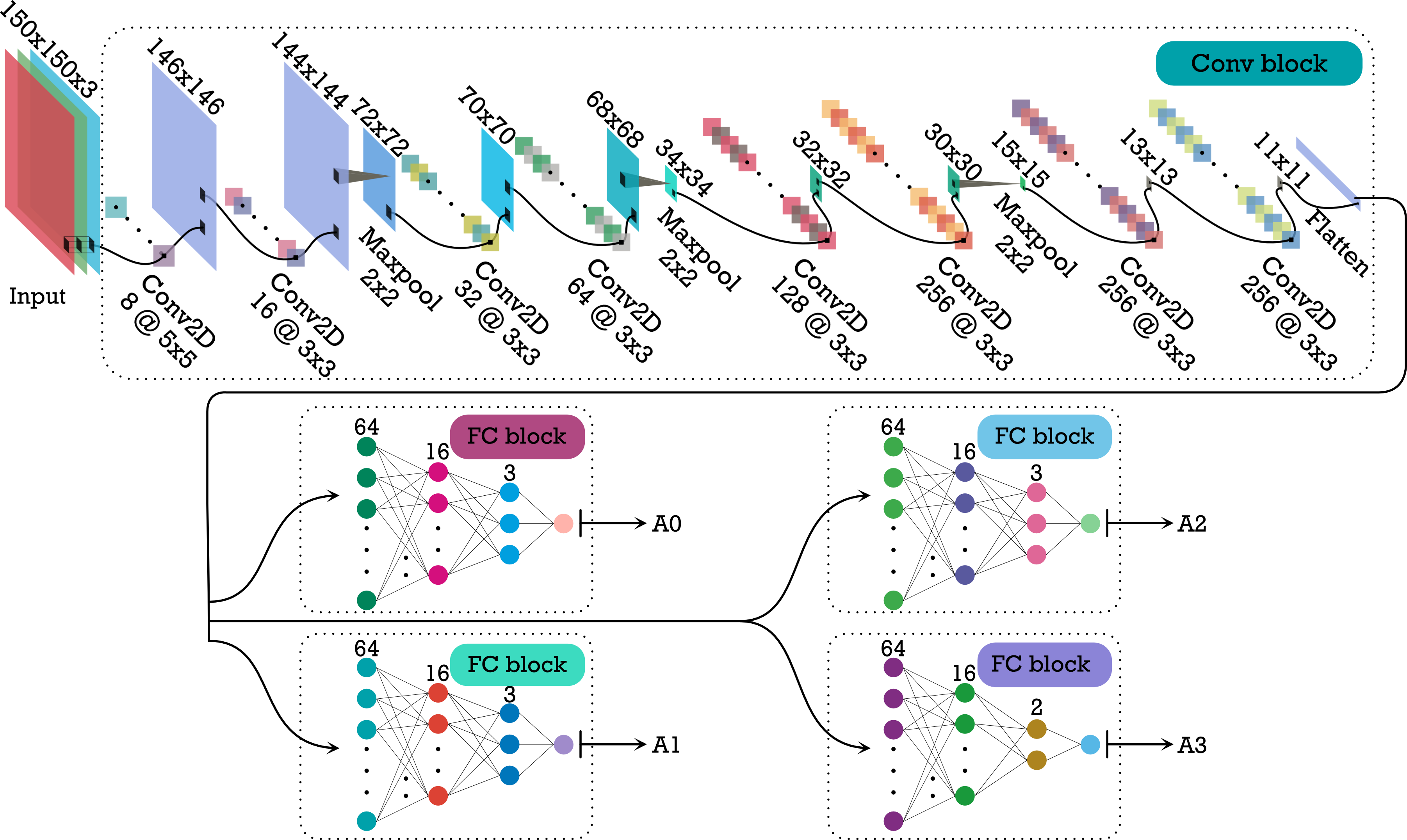}
      \caption{The neural network layers for the pick-and-place task where actions are sampled independently.}
      \label{fig:independent_layers}
\end{figure}

Suppose that the action space is N-dimensional, then the joint distribution of the action $\mathbf{a}$ given the observation $\mathbf{o}$ is $p(\mathbf{a}|\mathbf{o}) = p(a_1, a_2, a_3, ..., a_N|\mathbf{o})$. Policy networks commonly make the assumption that the actions are independent when conditioned on the observation\cite{vrteleop}\cite{daml}\cite{mdn}. Under this assumption,

\begin{equation}
    p(\mathbf{a} | \mathbf{o}) = \prod_{i=1}^{N} p(a_i|\mathbf{o})
\end{equation}

With this assumption, each of the action probabilities may be modeled separately. Usually, policy networks construct a common intermediate representation and derive the individual action probabilities from the intermediate representation (Fig.~\ref{fig:independent_conv}). 
\begin{equation}
    \mathbf{f} = \pi_f (\mathbf{o})
\end{equation}
\begin{equation}
    a_i = \pi_{a_i} (\mathbf{f}),
\end{equation}
for $1 \leq i \leq N$.

However, the independence assumption does not always hold (Fig.~\ref{fig:intro_reach}). One way to address the problem is to not make the independence assumption and to directly model the joint probability. However, the set of possible outcomes in the joint action space increases exponentially with the dimensionality of the action space. For example, if each of the actions is categorical with 5 possible values, then the joint action space will have 625 possible values for 4-dimensional action space. Because of this, it may be infeasible to directly model the joint action space. In the next section, we consider alternative ways of modeling the joint action space without making the strong independence assumption.

\section{STOCHASTIC ACTION GENERATION}

We consider three different ways by which stochastic actions can be generated by a policy network.

\subsection{Autoregressive action generation}

\begin{figure}[!t]
      \centering
      \includegraphics[width=0.9\linewidth]{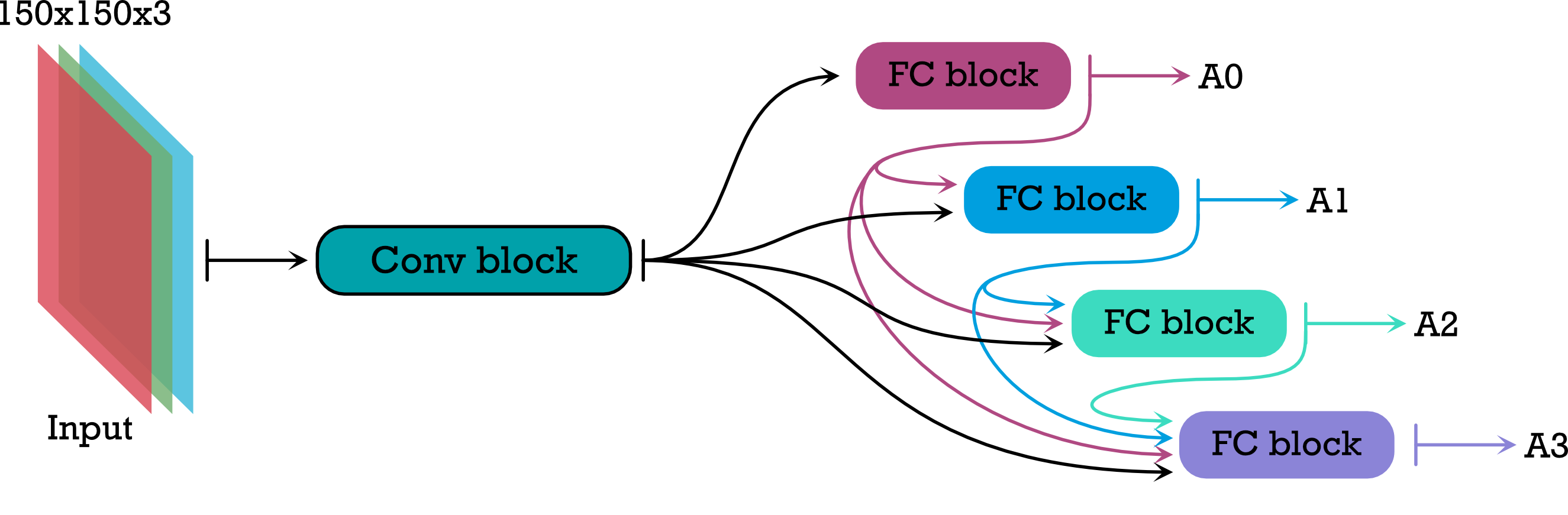}
      \caption{Autoregressive action prediction.}
      \label{fig:pixelcnn_conv}
\end{figure}

In autoregressive action generation, the joint distribution of the action $\mathbf{a}$ given the observation $\mathbf{o}$ is modeled as

\begin{equation}
    p(\mathbf{a}|\mathbf{o}) = \prod_{i=1}^{N} p(a_i | a_1, a_2, ..., a_{i-1}, \mathbf{o})
\end{equation}

The ordering of the action space may be arbitrarily chosen. With the order fixed, the distribution of each action depends on the actions before it and not on any other actions (Fig.~\ref{fig:pixelcnn_conv}). During training, the part of the policy network that outputs $a_i$ is fed expert actions in the dataset $\hat{a}_1, \hat{a}_2, ..., \hat{a}_{i-1}$ corresponding to the same observation. The outputs of the actions $a_1, a_2, ..., a_{i-1}$ are not sampled from the policy network during training. This is also sometimes referred to as ``teacher forcing". During inference, the first action $a_1$ is sampled unconditionally and then fed to the part of the policy network that generates $a_2$ and so on. For the example in Fig.~\ref{fig:intro_reach}, if $RIGHT$ is sampled as the first action, then the network will learn to predict $HOLD$ with a high probability for the second action.

One of the benefits of autoregressive action generation is that the training is stable. Unlike GANs or variational predictors, the training process has fewer problems such as mode collapse or difficulty in converging. However, during inference, since earlier actions are necessary to predict the subsequent actions, the inference is slow and inference time grows with the dimensionality of the action space. For action generation, unlike image generation, the dimensionality of the action space for most robots is often not too large and thus the inference time is manageable.

\subsection{Generative adversarial networks}

\begin{figure}[!t]
      \centering
      \includegraphics[width=0.9\linewidth]{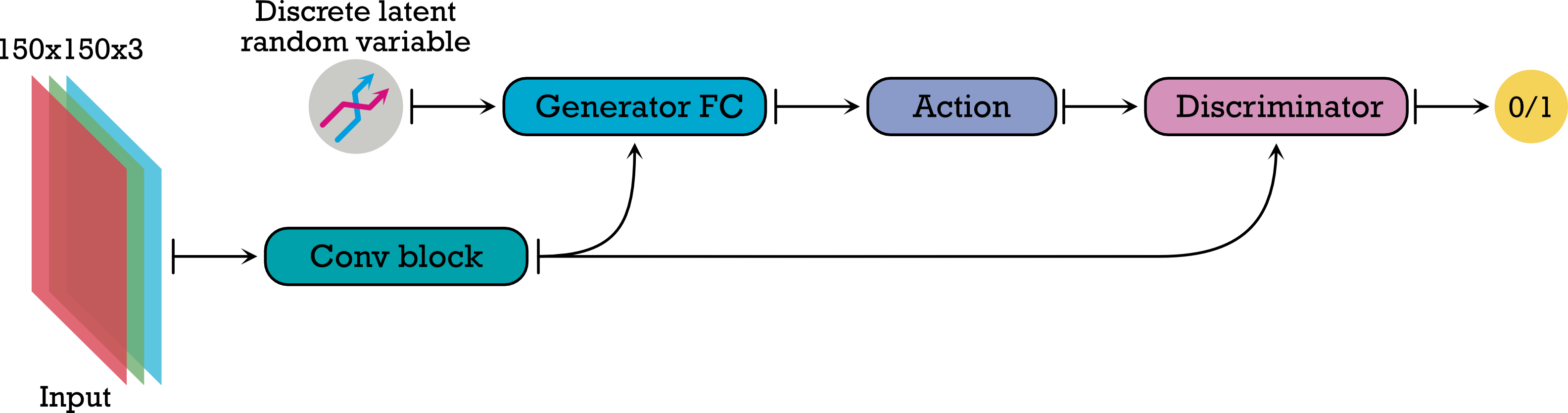}
      \caption{Generative adversarial network for action prediction.}
      \label{fig:gan_conv}
\end{figure}

In adversarial nets, a generator model is pitted against a discriminator (Fig.~\ref{fig:gan_conv}) that learns to predict whether a sample is from the distribution of the generator model or the data distribution\cite{gan}. The generator model $G$ parameterized by $\theta_g$ maps a prior noise distribution $p_{\mathbf{z}}(\mathbf{z})$ to the generator distribution $p_g$ as $G(\mathbf{z}; \theta_g)$. The discriminator model $D$ parameterized by $\theta_d$ predicts whether a sample $\mathbf{x}$ is sampled from the data distribution or the generator distribution. While the generator is trained to confound the discriminator, the discriminator is simultaneously trained to distinguish between the generator distribution and the data distribution.

\begin{equation}
    \begin{split}
        \min_{\theta_g} \max_{\theta_d} V(G, D) &= \mathbb{E}_{\mathbf{x} \sim p_{data}(\mathbf{x})}[\log (D(x))] \\
        & + \mathbb{E}_{\mathbf{z} \sim p_{\mathbf{z}}(\mathbf{z})}[\log (1-D(G(\mathbf{z})))]
    \end{split}
\end{equation}

In the conditional variant of GANs, both the generator and discriminator are fed the data on which we wish to condition the GAN\cite{cgan}. While GANs have been widely used for natural image generator, they can also be used for policy networks in imitation learning by conditioning the GAN on the observation $\mathbf{o}$. This will induce the generator to learn the expert data distribution conditioned on the observation $\mathbf{o}$. For the example in Fig.~\ref{fig:intro_reach}, the generator will learn to not emit $(RIGHT, DOWN)$ since the discriminator can easily predict that it is not in the training data.

A major problem with GANs is mode collapse where the generator exploits a local minima in the discriminator and learns to produce only a part of the data distribution which is enough to fool the discriminator\cite{wgan}. For policy networks generating actions, this problem is exacerbated because it can be difficult to recognize whether mode collapse has occurred whereas with natural images, visual inspection can be revealing. Furthermore, the consequence of mode collapse is also difficult to predict. As we show in experimental results, in some cases mode collapse does not harm the performance of the robot in performing the task, whereas for other tasks, it can result in catastrophic failure.

Unlike the autoregressive generator, the generator in GANs are efficient to sample from on modern GPUs as there is no need to serially sample one action after another.

\subsection{Variational action prediction}

\begin{figure}[!t]
      \centering
      \includegraphics[width=0.9\linewidth]{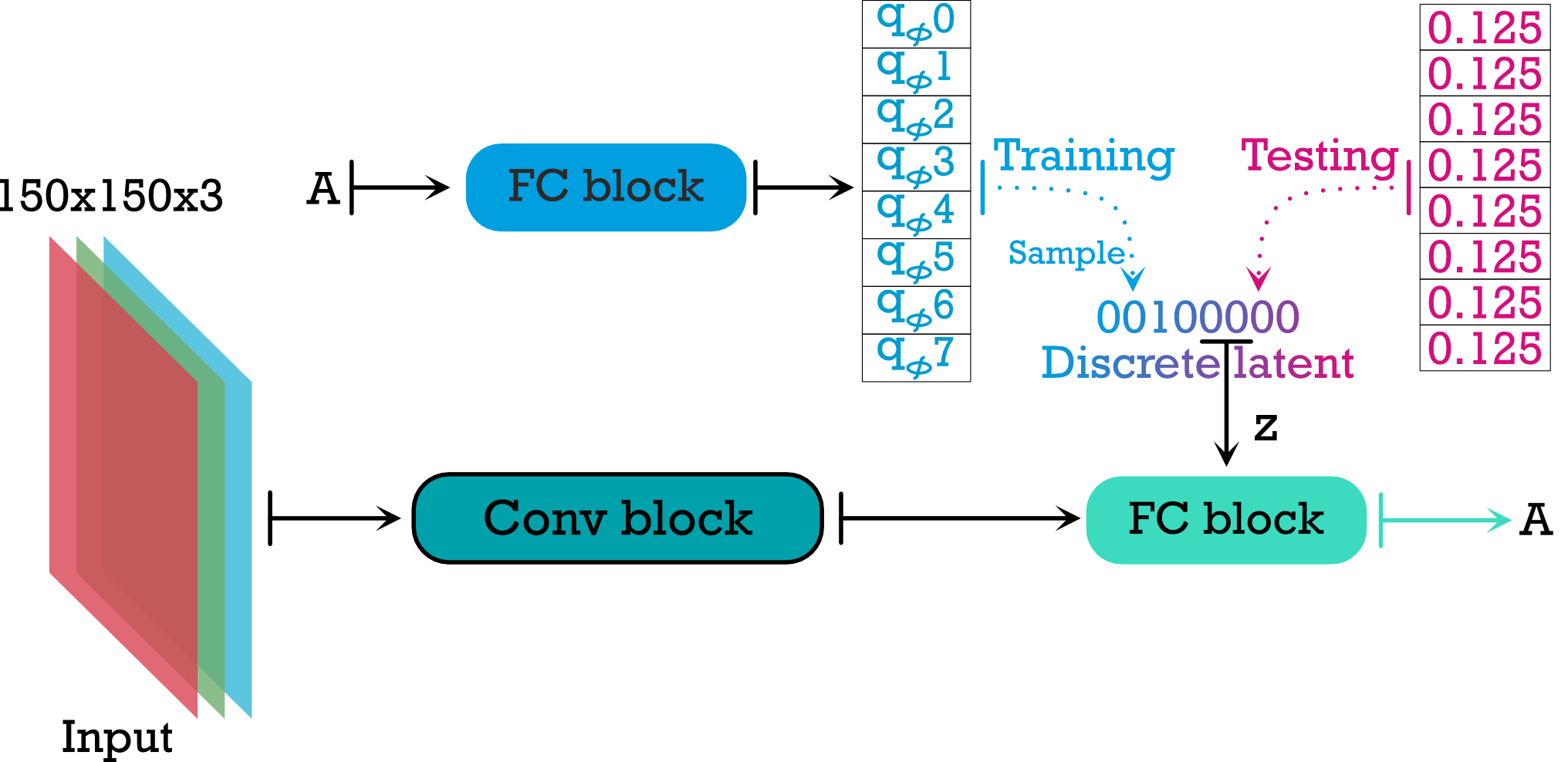}
      \caption{Variational action prediction.}
      \label{fig:variational_conv}
\end{figure}

Action prediction can be thought of as being stochastic due to latent events that are not present in the observation. In variational action prediction, a latent variable $\mathbf{z}$ that is drawn from the prior $\mathbf{z} \sim p(\mathbf{z})$ is introduced to build a model $p_\theta(\mathbf{a}|\mathbf{o},\mathbf{z})$ parameterized by $\theta$ (see Fig.~\ref{fig:variational_conv}). The true posterior $p(\mathbf{z}|\mathbf{a})$ is approximated with a network $q_{\phi}(\mathbf{z}|\mathbf{a})$. If this network predicts the parameters of a conditional Gaussian distribution $\mathcal{N}(\mu_\phi(\mathbf{o}), \sigma_\phi(\mathbf{o}))$, it can be trained using the reparameterization trick\cite{variational}

\begin{equation}
    \mathbf{z} = \mu_\phi(\mathbf{o}) + \sigma_\phi(\mathbf{o}) \times \epsilon, \quad \epsilon \sim \mathcal{N}(\mathbf{0}, \mathbf{I})
\end{equation}

Alternatively, if the prior $p(\mathbf{z})$ is a uniform distribution of discrete categorical random variables, the network $q_{\phi}(\mathbf{z}|\mathbf{a})$ that predicts the parameters of the distribution can be trained using the gumbel-softmax reparameterization trick\cite{concrete}\cite{gumbelsoftmax}. We use this latter  distribution in our experiments.

To learn the parameters $\theta$ and $\phi$, the variational lower bound\cite{variational} is optimized. The loss function is

\begin{equation}
    \label{eq:variational}
    \begin{split}
    L(\mathbf{o}, \mathbf{a}) = & - \mathbb{E}_{q_\phi(\mathbf{z}|\mathbf{a})} [\log p_\theta (\mathbf{a} | \mathbf{o}, \mathbf{z})]\\
    & + D_{KL}(q_\phi(\mathbf{z}|\mathbf{a})||p(\mathbf{z}))
    \end{split}
\end{equation}

The first term in the RHS of Eq.~\ref{eq:variational} is the action prediction loss and the second term is the divergence of the variational posterior from the prior on the latent variable.

During test time, the latent variable $\mathbf{z}$ is sampled from the assumed prior $p(\mathbf{z})$. Like GANs, action prediction with variational prediction is fast since all the actions may be simultaneously predicted. However, an additional hyper-parameter that scales the KL divergence in the loss function is introduced during training. In practice, the outcome of training is sensitive to the weight term associated with the KL divergence loss term\cite{modelatari}. 

\section{RESULTS}

We have two different experimental setups to investigate stochasticity in expert demonstrations and its impact on imitation learning. One is a toy remote-controlled (RC) car and the other is with a robotic arm. In all cases, the network is trained using the Adam optimizer\cite{adam}.

\subsection{Remote controlled Car}

\begin{figure}[!t]
    \centering
    \begin{subfigure}{.4\linewidth}
        \centering
        \includegraphics[width=.9\linewidth]{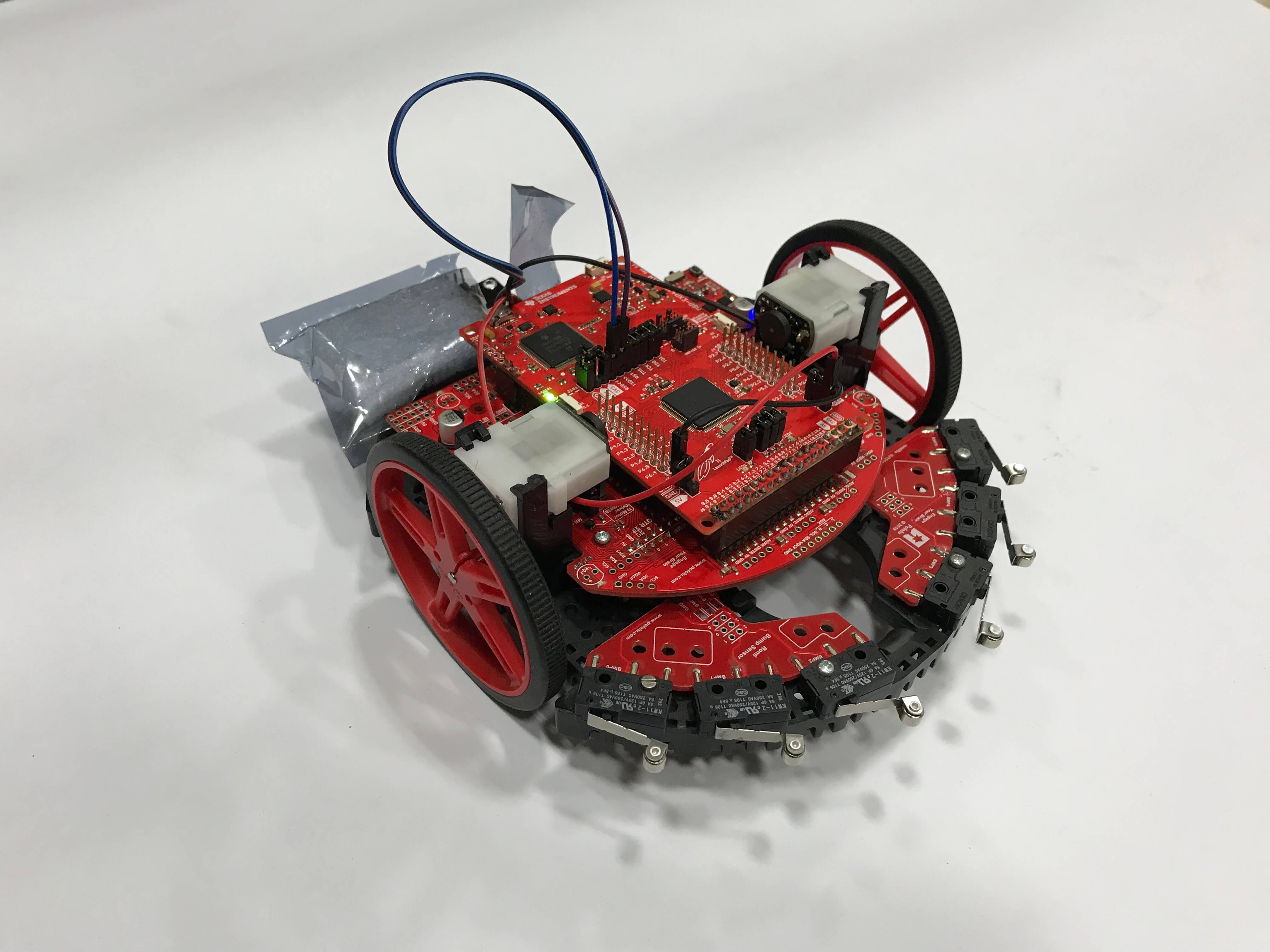}
        \label{fig:ti_rc_sub1}
    \end{subfigure}%
    \begin{subfigure}{.4\linewidth}
        \centering
        \includegraphics[width=.9\linewidth]{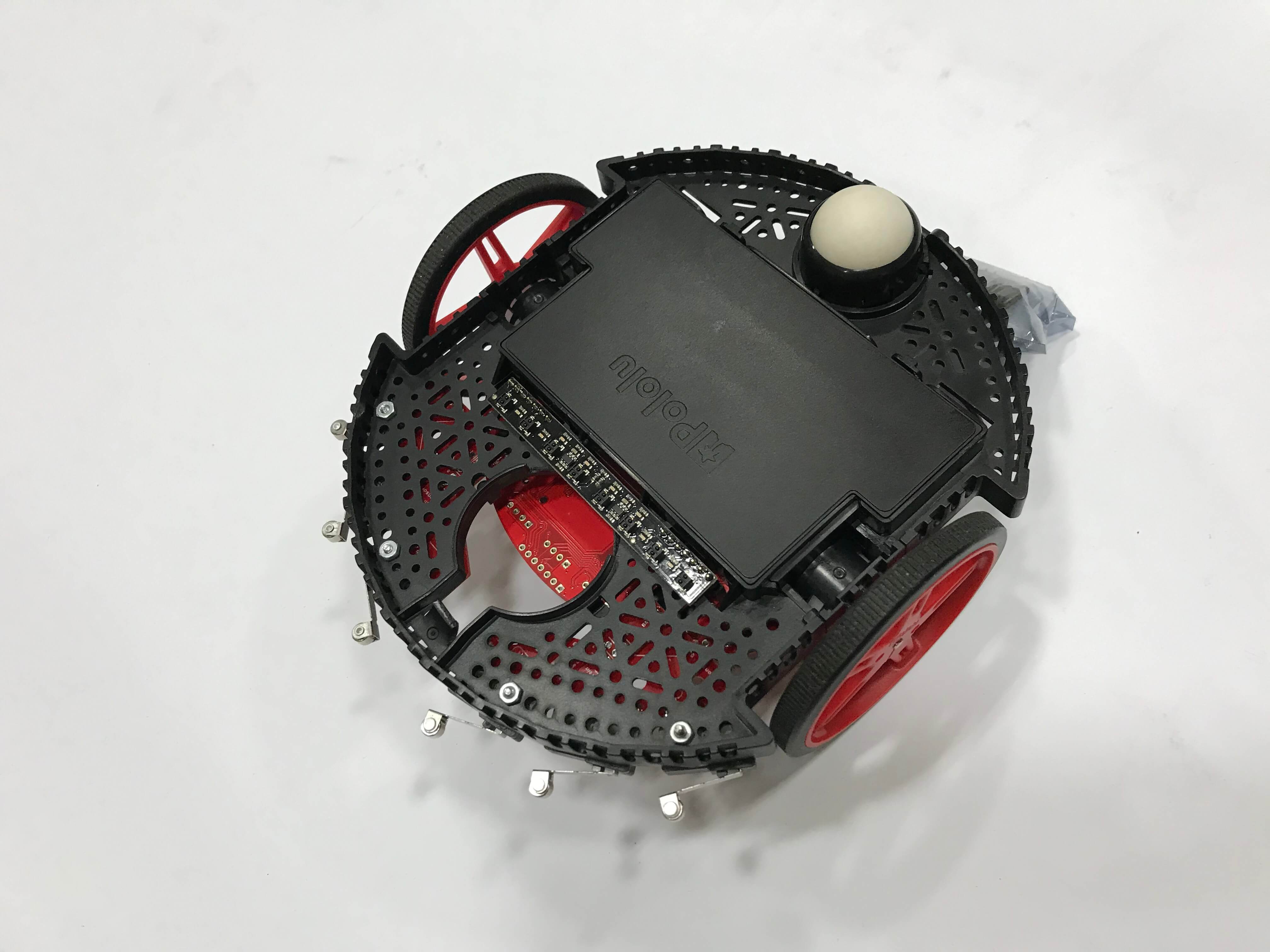}
        \label{fig:ti_rc_sub2}
    \end{subfigure}
    \caption{The TI remote controlled car}
    \label{fig:ti_rc}
\end{figure}

The RC car based on Texas Instruments Robot System Learning Kit is shown in Fig.~\ref{fig:ti_rc}. It has two motors driving the two wheels which are independently controlled through pulse width modulation (PWM). An ESP8266 module is used for remote control via WiFi. Commands to the motors are sent via UDP at 5~Hz. The torque applied to the motors using PWM can be controlled by expert demonstrators using the analog triggers on an Xbox 360 game controller. The car also has eight photoreceptors at the bottom along with IR emitters (Fig.~\ref{fig:ti_rc}) which captures an 8-bit binary ``image" where a bit is set to 1 depending on whether there is a dark line right below the photo-diode. For each experiment, 10 expert demonstrations are collected for training, and the trained network is evaluated 10 times. The details of the network architecture are in the appendix.

\subsubsection{Driving straight}

\begin{figure}[!t]
    \centering
    \begin{subfigure}{.49\linewidth}
        \centering
        \includegraphics[width=1\linewidth]{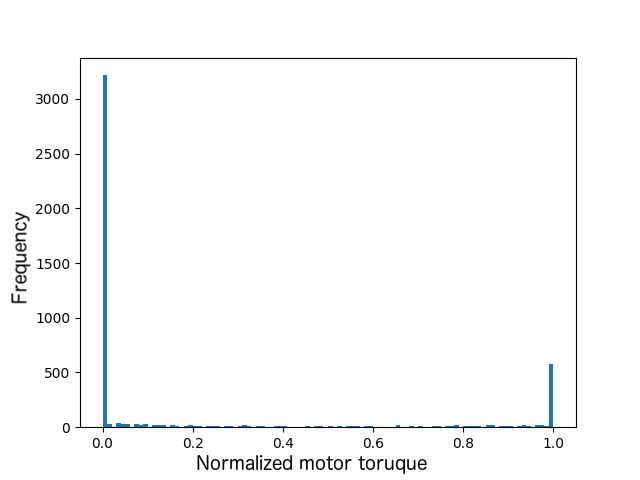}
        \label{fig:hist_sub1}
    \end{subfigure}%
    \begin{subfigure}{.49\linewidth}
        \centering
        \includegraphics[width=1\linewidth]{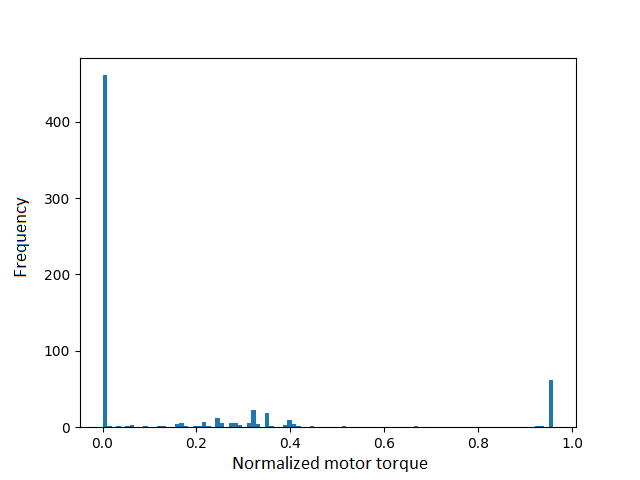}
        \label{fig:hist_sub2}
    \end{subfigure}
    \caption{Histogram of normalized PWM values in the training data for the left and right wheel respectively.}
    \label{fig:hist}
\end{figure}

In the first experiment, the goal was to simply drive straight. The reason why this is not trivial is that the two motors are not identical, and they do not provide any feedback of the torque or velocity. As a result, the demonstrator has to observe the path being taken by the car and adjust the commanded PWM using the analog triggers on the game controllers so that the car moves forward on a straight line. Figure~\ref{fig:hist} shows the distribution of the left and right PWM values collected from expert demonstrations. %Furthermore, since the response of the motors is non-linear, the precise ratio of the PWM value set for the left motor to the right motor depends on the absolute velocity of the car. 

We define a trajectory as being successful if the car moves forward by 8 feet while staying within 30~cm of the straight-line trajectory. As a baseline, we use a Gaussian mixture model (GMM) with 4 Gaussians. This is compared against the other considered methods in Table~\ref{table:straight}.% The autoregressive action generation performs better than independently sampling the left and right motor PWMs using a GMM.

\begin{table}[!t]
    \centering
    \begin{tabular}{l l} 
        \hline
        Method & Success rate \\ [0.5ex] 
        \hline
        Independent sampling & 60\% \\
        Autoregressive sampling & 90\% \\
        GAN & 70\% \\
        Variational prediction & 90\% \\
        \hline
    \end{tabular}
    \caption{Success rate in driving the RC car straight for 10 trials.}
    \label{table:straight}
\end{table}

\subsubsection{Line following}

\begin{figure}[!t]
      \centering
      \includegraphics[width=0.7\linewidth]{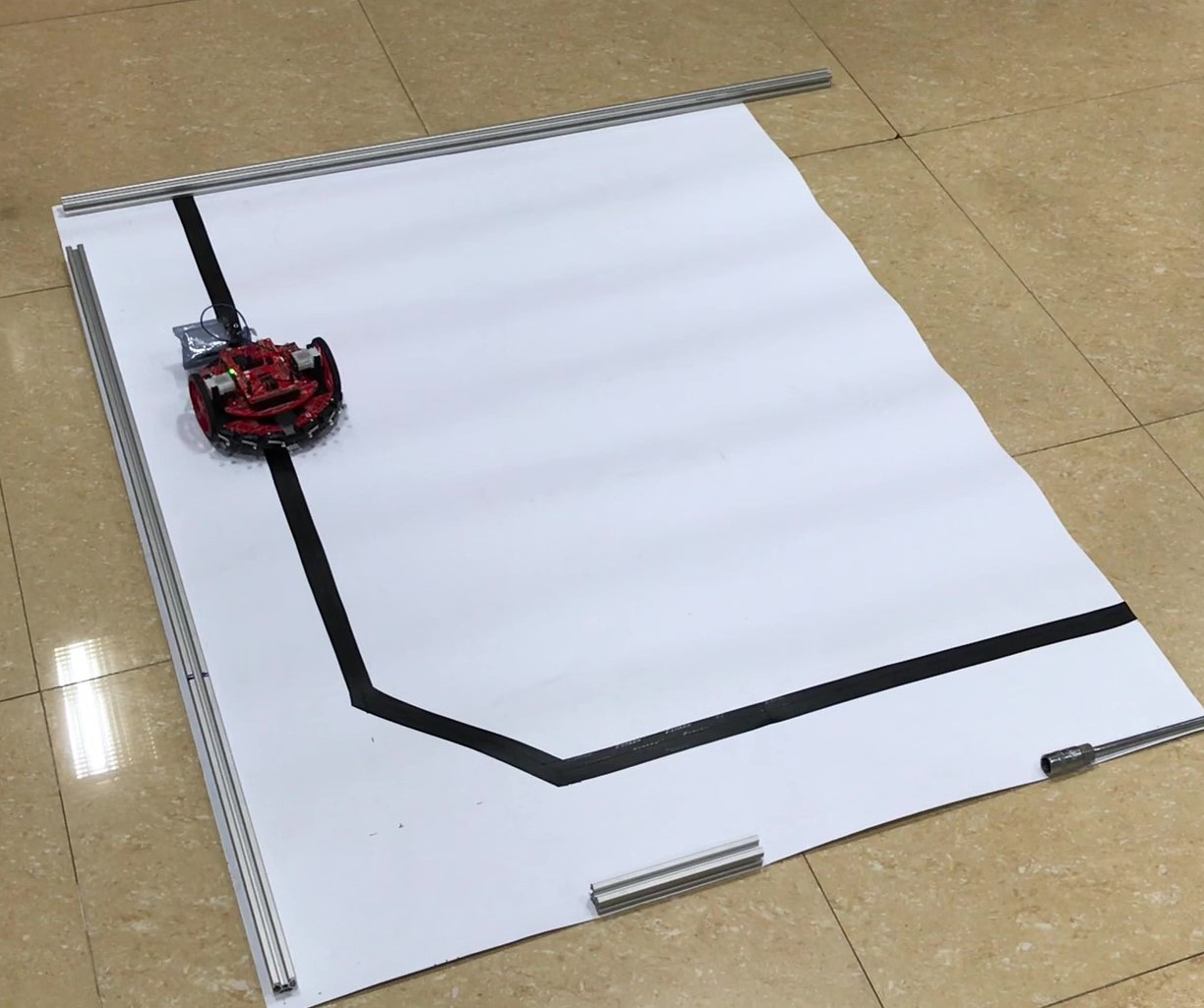}
      \caption{The RC car following a line.}
      \label{fig:line_following}
\end{figure}

The second task with the RC car is line following. When a photosensor is just above the black line in Fig.~\ref{fig:line_following}, the sensor measures a ``1". So, when the center of the car is above the line, the 8-bit observation is typically ``00011000". Likewise, when the car has to turn left because it is moving to the right of the line, the observation might be ``11000000". In some demonstrations, the course correction is swift and happens on even small deviations when the observation might be ``00110000", but in other demonstrations, it is delayed until the car reaches the edge of the line (``11000000"). Similarly, some demonstrations include turning at high speeds, whereas others are slow. These are sources of stochasticity. If the velocity of the left and right wheels are sampled independently, we can expect frequent errors which consume additional time to correct. The duration it takes to complete the course is detailed in Table~\ref{table:line_following}. We see that modeling stochasticity results in faster completion times. However, for both tasks, GANs performed worse. We noticed that while the generator successfully modeled the stochasticity in the dataset, the output of the generator was biased and consistently gave out higher PWMs for one of the wheels which resulted in inferior performance.

\begin{table}[!t]
    \centering
    \begin{tabular}{l l} 
        \hline
        Method & Duration (s) \\ [0.5ex] 
        \hline
        Independent sampling & 90.28 \\
        Autoregressive sampling & 70.28 \\
        GAN & 85.1 \\
        Variational prediction & 66.9 \\
        \hline
    \end{tabular}
    \caption{Average time taken to complete the line following task over 10 trials.}
    \label{table:line_following}
\end{table}

\subsection{Robot arm}

%\begin{figure}[!t]
%      \centering
%      \includegraphics[width=0.75\linewidth]{dobot_motomini_crushed.jpg}
%      \caption{The Dobot Magician and the Yaskawa MotoMini robot arms.}
%      \label{fig:dobot_motomini}
%\end{figure}

\begin{table}[!t]
    \centering
    \begin{tabular}{l l l l} 
        \hline
        \multirow{2}{*}{Method} &
        \multicolumn{3}{c}{Success rate} \\
        & Reach & Push & Pick-and-place \\
        \hline
        Independent sampling & 60\% & 26.6\% & 20\% \\
        Autoregressive sampling & 100\% & 100\% & 100\% \\
        GAN & 100\% & 0\% & 0\% \\
        Variational prediction & 100\% & 100\% & 100\% \\
        \hline
    \end{tabular}
    \caption{Success rate with the robotic arm for different tasks.}
    \label{table:arm}
\end{table}

We use the Yaskawa MotoMini (Fig.~\ref{fig:push}) and the Dobot Magician (Fig.~\ref{fig:pnp_gan}) for manipulation tasks. The observations are 150$\times$150 pixel images from a camera above the robot along with the current position of the end-effector and the gripper open/close state. The action space is 4-dimensional and includes the 3D direction of movement of the end-effector at every time step and is discretized to $\{0, +1, -1\}$. It also includes a binary gripper open/close command. The camera is sampled and end effector movement instructions are issued to the robot at 5 frames per second. For each task, the network is trained with 15 expert demonstrations and evaluated over 15 trials. The performance of different approaches is compared in Table~\ref{table:arm}.

\subsubsection{Reach task}
The reaching task is performed with the Yaskawa MotoMini (Fig.~\ref{fig:intro_reach}). The goal is to reach the target ``X" while avoiding obstacles. We consider the task to be successfully completed if the end-effector is brought within 2~cm of the target while not touching the obstacle. The stochasticity arises because on encountering an obstacle, experts behave differently; some move around the obstacle and some above it. With GANs, we observed mode collapse which caused the robot to always turn right when it encountered an obstacle. However, mode collapse does not impede the completion of the task.

\subsubsection{Pushing a pen}

\begin{figure}[!t]
    \centering
    \begin{subfigure}{.4\linewidth}
        \centering
        \includegraphics[width=.9\linewidth]{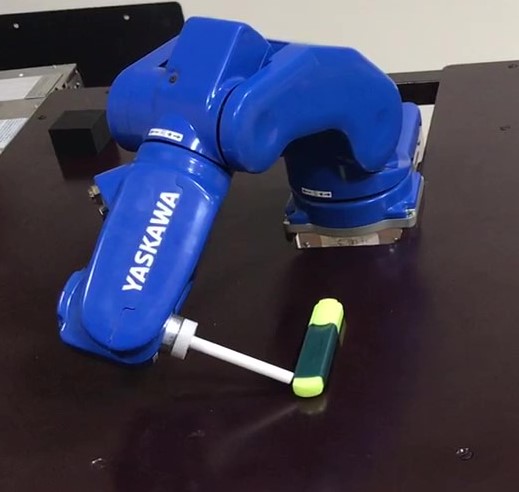}
    \end{subfigure}%
    \begin{subfigure}{.4\linewidth}
        \centering
        \includegraphics[width=.9\linewidth]{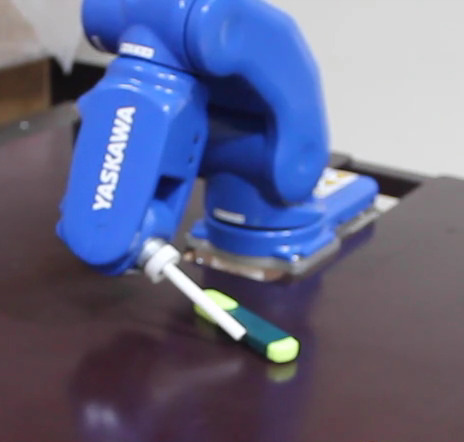}
    \end{subfigure}
    \caption{The robot pushes the pen towards the edge of the table. If the action space is independently sampled, the robot can push the pen to a horizontal position (parallel to the end-effector tool) from which it cannot recover.}
    \label{fig:push}
\end{figure}

The push task is also performed using the MotoMini. The task is to push a pen on a table. We consider the task to be successfully completed if the pen is moved by at least 30~cm across the table (Fig~\ref{fig:push}). We noticed stochasticity in the expert demonstrations when the end-effector is close to the tip of the pen. Some of the expert demonstrations involve pushing the pen forward; in others, the end-effector is moved towards the other end of the pen. If the different dimensions of the action space are sampled independently, the robot can push the pen to a horizontal position (Fig~\ref{fig:push}). If that happens, the observations are now outside the training distribution since this scenario never occurs in the training data, and the robot does not recover. While autoregressive action generation addresses this issue, GANs perform poorly due to mode collapse which causes the end effector to oscillate about the pen, and the robot does not push the pen forward at all. While this particular set of actions can confuse the discriminator, it does not capture all the modes in the training data and leads to complete failure in completing the task. We also tried training the Wasserstein GAN\cite{wgan} to address this problem, but even with WGAN, the mode collapse was present.

\subsubsection{Pick and Place}

\begin{figure}[!t]
    \centering
    \begin{subfigure}{.49\linewidth}
        \centering
        \includegraphics[width=.9\linewidth]{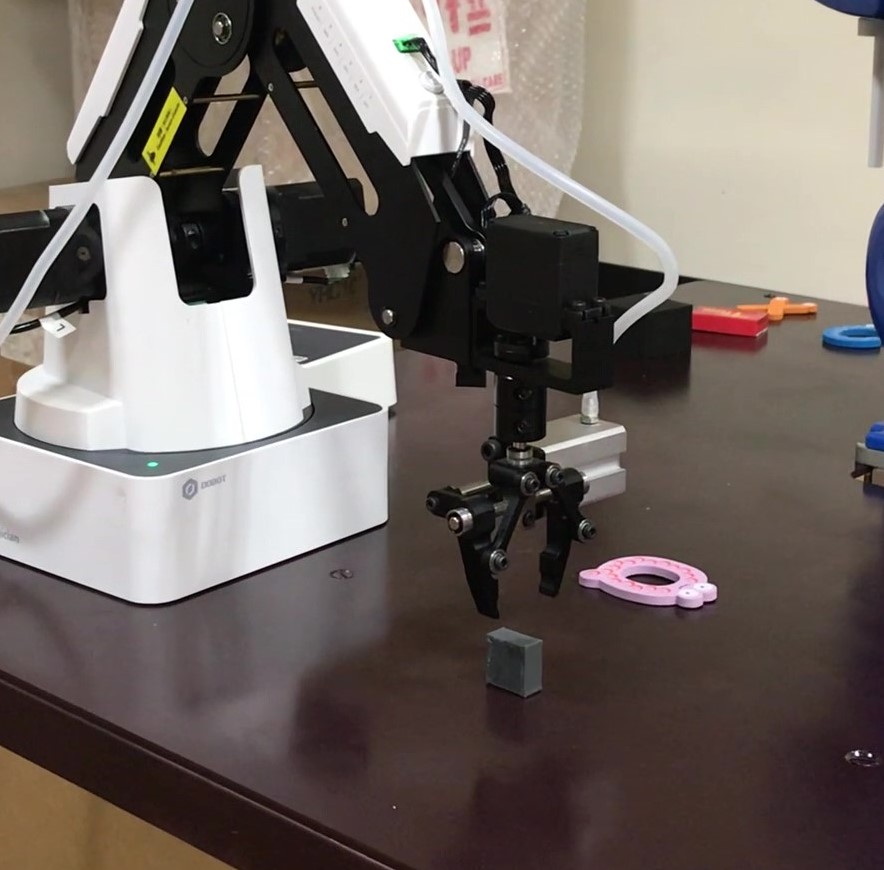}
    \end{subfigure}%
    \begin{subfigure}{.49\linewidth}
        \centering
        \includegraphics[width=.9\linewidth]{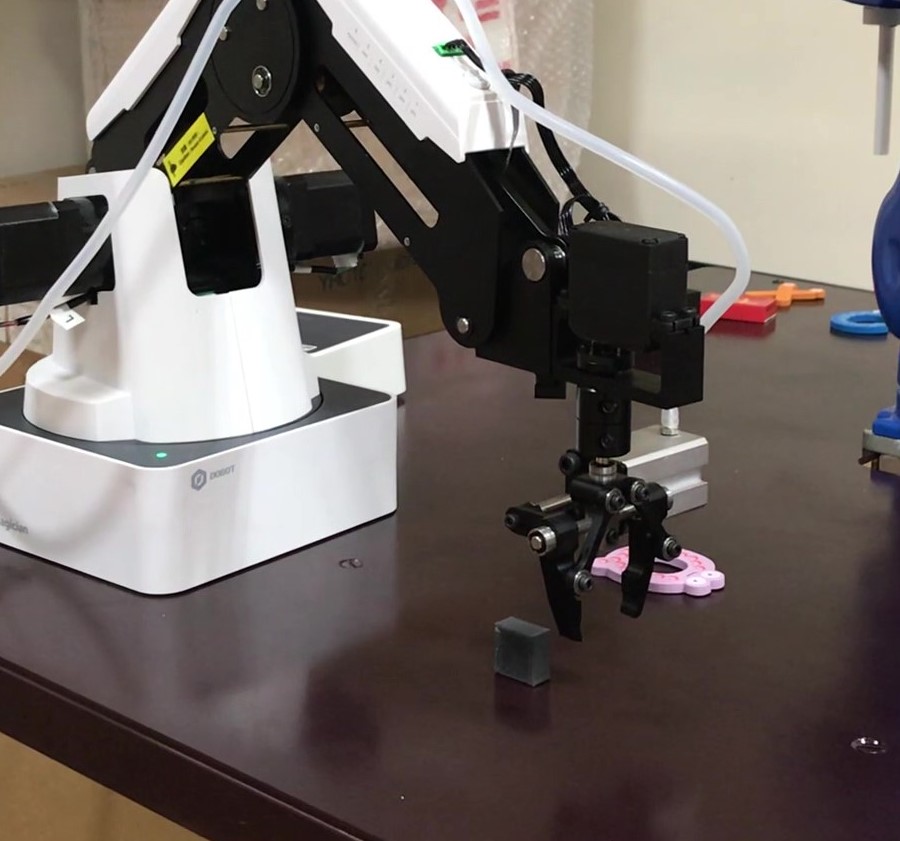}
    \end{subfigure}
    \caption{Mode collapse in GAN causes the end effector to oscillate around the object without picking it up.}
    \label{fig:pnp_gan}
\end{figure}

The pick-and-place task was performed with the Dobot Magician. We consider the task to be successful if the cuboid is placed inside the hollow part at the center of the target shaped ``O" in Fig.~\ref{fig:pnp_gan}. We observed that sometimes the demonstrator overshoots the object (or target) and corrects for this at the next timestep. So for similar observations, we have two different actions. If the action space is independently sampled, the robot attempts to grip the object at a position where the object is not present. Likewise, the object is sometimes dropped too far from the target because the act of moving and releasing the gripper are independently sampled. Autoregressive sampling addresses this problem effectively but GANs perform poorly due to mode collapse. Similar to the pushing task, the gripper approaches the object and overshoots, but due to mode collapse, the robot always overshoots and never picks up the object and oscillates around it, as shown in Fig.~\ref{fig:pnp_gan}.

In all our experiments, autoregressive action prediction successfully accounts for stochasticity in the training data and is the easiest to train since it does not introduce any new hyperparameters. However, since actions are sequentially sampled, the inference time is proportional to the dimensionality of the action space. GANs do not require such sequential sampling, but suffers from mode collapse. Variational prediction also offers fast inference since it does not require sequential sampling of the action space, but it does introduce additional hyper-parameters for training.

\section{CONCLUSIONS}

In this paper, we have described a teleoperation system from which one can collect expert demonstrations for simple tasks such as driving, pushing an object, and picking and placing an object. We have shown that such demonstrations exhibit stochasticity which can impede the performance of imitation learning if ignored. With multi-dimensional action space, sampling the different actions independently results in sub-optimal imitation learning performance. Autoregressive sampling, GANs, and variational predictors are three tractable ways to model stochasticity in the data. Autoregressive sampling is easiest to train but is slow during inference while mode collapse in GANs is a serious problem for imitation learning.

\addtolength{\textheight}{-12cm}   % This command serves to balance the column lengths

%%%%%%%%%%%%%%%%%%%%%%%%%%%%%%%%%%%%%%%%%%%%%%%%%%%%%%%%%%%%%%%%%%%%%%%%%%%%%%%%

%%%%%%%%%%%%%%%%%%%%%%%%%%%%%%%%%%%%%%%%%%%%%%%%%%%%%%%%%%%%%%%%%%%%%%%%%%%%%%%%

%%%%%%%%%%%%%%%%%%%%%%%%%%%%%%%%%%%%%%%%%%%%%%%%%%%%%%%%%%%%%%%%%%%%%%%%%%%%%%%%

\section*{APPENDIX}

For the line following task, instead of conv layers, a feature vector is derived by passing the 8-bit input ``image" through FC128-ReLU-FC16. This is used to obtain the parameters of a GMM (with 4 mixtures). For the push, the conv layers used are shown in Fig.~\ref{fig:independent_layers}, but the output FC layers are FC128-FC3.

\section*{ACKNOWLEDGMENT}

We would like to thank Ashish Joglekar for his assistance with the TI robot kit.

%%%%%%%%%%%%%%%%%%%%%%%%%%%%%%%%%%%%%%%%%%%%%%%%%%%%%%%%%%%%%%%%%%%%%%%%%%%%%%%%

\end{document}